\documentclass[conference]{IEEEtran}
\IEEEoverridecommandlockouts

\usepackage{cite}
\usepackage{amsmath,amssymb,amsfonts}
\usepackage{algorithmic}
\usepackage{graphicx}
\usepackage{caption}
\usepackage{subcaption}
\usepackage{subfig}
\usepackage{textcomp}
\usepackage{xcolor}
\def\BibTeX{{\rm B\kern-.05em{\sc i\kern-.025em b}\kern-.08em
    T\kern-.1667em\lower.7ex\hbox{E}\kern-.125emX}}
\begin{document}

\title{
A Multi-Dimensional Clustering Approach for
Identifying Inborn Errors of Immunity\\
\thanks{Funded in part by Immune Deficiency Foundation (IDF)
}
}
\author{\IEEEauthorblockN{Nishad Kulkarni, \textsuperscript{1*}, Alexandra K. Martinson,\textsuperscript{2,5*},}
\IEEEauthorblockN{Nicholas L. Rider,\textsuperscript{3}, Michael Keller,\textsuperscript{2,4,5}, Syed Muhammad Anwar,\textsuperscript{1,5}}
\\\textsuperscript{1}{Sheikh Zayed Institute for Pediatric Surgical Innovation, Children's National Hospital, Washington, DC 20010}
\\\textsuperscript{2}{Childrens National Hospital, Washington, DC}
\\\textsuperscript{3}{Department of Health Systems \& Implementation Science, Division of Allergy \& Immunology}
\\{Virginia Tech Carilion School of Medicine, Roanoke, VA}
\\\textsuperscript{4}{Division of Allergy \& Immunology Childrens National Hospital, Washington, DC}
\\\textsuperscript{5}{School of Medicine and Health Sciences, George Washington University, Washington, DC 20052}
\\
{\footnotesize \textsuperscript{*}These authors contributed equally to this work and are co-first authors}
}




\maketitle

\begin{abstract}
Rare diseases such as inborn errors of immunity (IEI) require early diagnosis to prevent end organ damage and improve quality of life. Hurdles in accessing and curating large scale electronic health record (EHR) data limit routine data driven analyses to remain on the forefront of IEI and other rare disease trends. 
Development of machine learning (ML) algorithms in IEI for pattern recognition as well as published methodology examining how to systematically process and integrate complex medical data is limited. Our proposed pipeline, including data curation and ML clustering algorithms, is designed to recognize novel rare disease patterns and extract IEI-associated features from a national data registry. Our methodology for EHR data formatting and processing presents the pipeline that transforms raw immunologic lab data into vectors. This is further combined with hyperparameter tuning for diseases pattern recognition via clustering. This study refines IEI feature awareness, develops data tool kits for rare disease populations analysis, and expands on transforming complex medical records in data structures interpretable by unsupervised ML. 
\end{abstract}


\section{Introduction}
The complexity of disease symptoms in patients with rare diseases such as Inborn Errors of Immunity (IEI) results in arriving at the correct diagnosis a prolonged saga across many healthcare visits. Machine Learning (ML) approaches have not yet been commonly applied to the detection or pattern recognition of IEI [1]. 
We demonstrate the application of unsupervised computational algorithms using laboratory data as a proof of concept for rare disease pattern analysis using a national health data warehouse with large volumes of IEI related electronic health records (EHR) [2]. Hence, our overarching goal is to identify novel disease patterns within IEI.

\noindent \paragraph*{\textbf{Related Works}} Prior IEI screening guidelines emphasize the need for pattern recognition of predisposing medical and demographic factors associated with IEI [3-4]. Patients may present over years with seemingly unrelated disorders across health centers. With the rapidly evolving nature of IEIs, it is vital to continuously analyze data for these patients. 
The current use of clustering, similar to pattern recognition, algorithms has been widely studied on medical imaging with a minority focused on electrocardiogram, flow cytometry, and genomics/proteomics [5-9]. A limited number of groups have worked on ML disease prediction algorithms for IEI diagnosis [8, 10-13]. 
ML clustering algorithms for pattern recognition in IEI have only focused on flow cytometry immunophenotyping data in a categorical format (low vs. high relative to normal values) [8]. Clustering models on laboratory data beyond flow cytometry have been published infrequently and focused on well known diseases with imputation using the entire dataset and low number of target clusters [14-15]. The use of clustering algorithms for rare disease pattern recognition and subphenotyping on multi-center lab data using disease-based imputation has not been well studied.

The goal of our proposed clustering pipeline for IEI pattern recognition is to demonstrate a methodology for transforming complex medical data into clinically interpretable data structures, while taking into account data missingness, normalization of data variety across clinical centers and lab values, and dimensionality reduction. Clustering is performed on measurements of immune system cells that protect against infections such as pneumonia and sepsis, and takes into account age-based subgroup analysis necessary for building algorithms applicable to pediatric populations with rapidly changing immune systems [16-18]. By analyzing patterns in these laboratory tests, we aim to predict IEI subcategories, or sub-phenotypes, not currently labeled in our data. 

\begin{figure*}[h]
\centering
\includegraphics[width=0.75\linewidth]{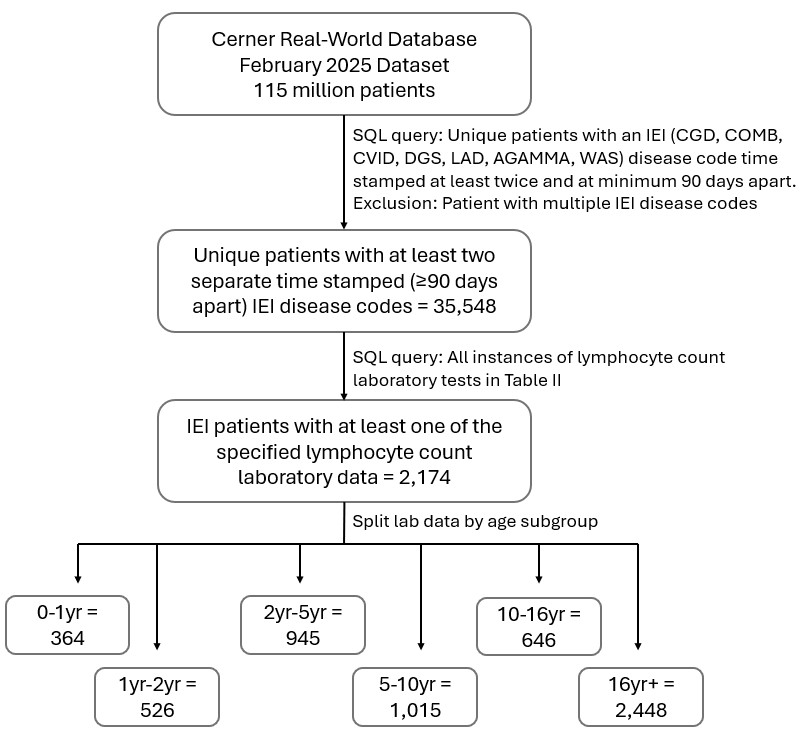}
\caption{Study population and query criteria.}
\label{fig:data structure}
\end{figure*}

Our main contributions are: 
\begin{enumerate}
    \item Demonstration of detailed IEI related data processing steps necessary for clinical data representation into a format interpretable by clustering algorithms.
    \item Leveraging centroid-based, density-based, and hierarchical ML clustering algorithms on laboratory data for patients with presumed IEI.
    \item Use of unsupervised clustering algorithms to identify IEI disease subgroups that can be further evaluated for new disease subtypes.
\end{enumerate}

\section{Methodology}
\color{black} Our proposed methodology to identify clusters related to IEI is presented in the following subsections.  \color{black}


\subsubsection*{\textbf{Data Identification and Structuring}}

All the data are accessed through the Cerner Real-World Database (CRWD) [2].
The patient population for this study includes 5,986 patient lab samples from unique hospital encounters across 2,174 patients (details in Figure 1, Table 1). The patient populations were extracted using disease specific ICD codes for only one of seven (Table \ref{tab:patients}) IEIs with at least two time stamped disease specific codes separated by a minimum of 90 days. IEIs included in this study included CGD: Chronic Granulomatous Disease, COMB: Combined Immunodeficiency, CVID: Common Variable Immunodeficiency, DGS: DiGeorge Syndrome, LAD: Leukocyte Adhesion Deficiency, AGAMMA: Agammaglobulinemia, and WAS: Wiskott Aldrich Syndrome.

The structure of the data is elaborated in Figure \ref{fig:data structure}. 
Each individual patient has a single patient ID, one or more encounter IDs, and one or more lab-date IDs. An encounter is when a patient checks into a hospital and may stay there over multiple hours/days. During the \textit{'encounter'}, any set of clinical measurements are performed over one or multiple dates. For instance, a patient could be tested for immune system function (i.e., CD3+ cells, CD19+ cells, etc.) at one time, and might get tested for other labs (i.e., CD3+CD4+ and CD19+) at another date/time. Hence, we end up with different number of occurrences of each clinical test for the given encounter. 

\begin{table}[!h]
    \centering
    \caption{Unique number of laboratory samples and patients for various IEI diagnoses considered in our curated data.}
    \begin{tabular}{|l|p{2.5cm}|}
        \hline
        \textbf{IEI Type} & \textbf{Lab Encounters (Unique Patients)}\\
        \hline
        \textbf{\textit{CGD}} - Chronic Granulomatous Disease& 314 (102)\\
        \hline
        \textbf{\textit{COMB}} - Combined Immunodeficiency& 664 (157)\\
        \hline
        \textbf{\textit{CVID}} - Common Variable Immunodeficiency& 305 (189)\\
        \hline
        \textbf{\textit{DGS}} - DiGeorge Syndrome& 1237 (753)\\
        \hline
        \textbf{\textit{LAD}} - Leukocyte Adhesion Deficiency& 30 (9)\\
        \hline
        \textbf{\textit{AGAMMA}} - Agammaglobulinemia& 3256 (930)\\
        \hline
        \textbf{\textit{WAS}} - Wiskott Aldrich Syndrome& 138 (34)\\
        \hline
    \end{tabular}
    \label{tab:patients}
\end{table}
It is important to note that longer hospital stays can include dozens of laboratory measurements for a patient. To simplify data and portray a progression over a patient's disease course as opposed to just one clinical encounter, we consider the first occurrence of each available test over a patient's hospital stay. Thus, each individual sample for clustering consists of an array containing the first recorded instance of each laboratory test during a hospital visit. 

Only patients that did not have overlapping IEI disease labels were included in this study. The data were divided based on patient age into 6 groups: 0-1 year, 1-2 years, 2-5 years, 5-10 years, 10-16 years and ${>}$ 16 years of age with the maximum age being the non-inclusive limit (Figure 1). The selected age subgroups are based on three major immunology studies detailing age-related changes in lymphocyte subpopulations, which informed the clustering data [16-18]. 

There is a reasonable probability of cases where a certain lab test is not available throughout the hospital visit. This can be due to isolated testing for specific lymphocytes or the more common problem of missing values in clinical data warehouses. For such cases we infer the missing values by applying the multiple imputation with multivariate imputation by chained equation (MICE) algorithm [19] on disease specific age-subgrouped laboratory data.


\begin{figure}[!t]
\centerline{\includegraphics[width=0.5\textwidth]{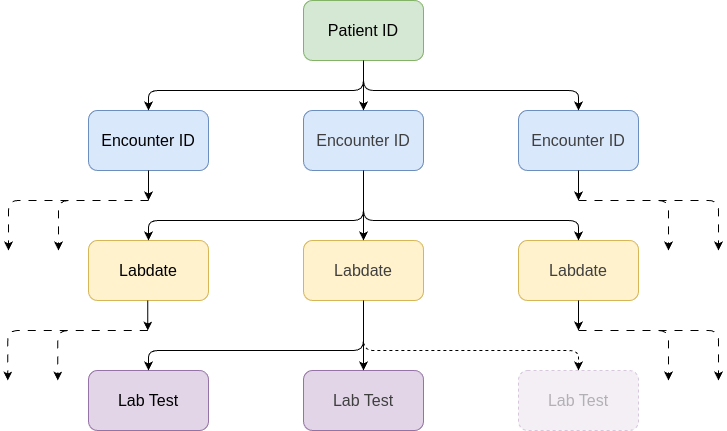}}
\caption{Data structure for EHR laboratory data for a unique patient ID.}
\label{fig:data structure}
\end{figure}

\subsubsection*{\textbf{Lab Tests}}

Each lab test has a reference high and reference low measurement associated with it as clinical data is based on a spectrum of normal results in the general population without disease. It is important to acknowledge in big data analysis across clinical sites that these test measurement reference ranges can vary and should be taken into account when interpreting laboratory values. We used each clinical site's reference range to normalize the laboratory measurement. 



The laboratory tests used in our algorithm were five clusters of differentiation (CD), types of receptor on our cells, relating to immune system function. Lab test \textit{'CD3+'} is a test of T lymphocytes, which express the CD3 receptor. The CD3+CD4+ and CD3+CD8+ labs test for combination receptors, representing helper and cytotoxic T lymphocytes respectively. 
Additionally, CD19+ samples were included and represent B cells, while CD3-,CD16+,CD56+ samples represents natural killer cells which kill harmful cells in the body. 
Since this is a pilot study to explore the feasibility of using clustering algorithms to predict phenotypes, we decided to limit the lab tests to the top 5 most clinically prominent tests. Additionally, this decision is also reinforced by the fact that more number of labs meant more missing data since all patients may not have the data for all the labs.

    ~ 
    
    \label{fig:combined}

We represent every data point as a vector consisting of five measurements across the lab ids. The following equation is for patient-encounter lab representation,
\begin{equation}
    \hat{L_t} = \frac{\sum_{i=1}^{n} a_i \hat{{l_i}}}{|\vec{L_t}|},
\end{equation}
where, $\vec{L_t}$ is the individual vector representation of a data point 
and $a_i \hat{\mathbf{l_i}}$ is the 
contribution of each lab test in the n-dimensional space. Normalization of lab components was calculated as, 
\begin{equation}
    \mathbf{L_i} = \begin{pmatrix}
        F_{Norm}(CD3+) \\
        F_{Norm}(CD3+, CD8+) \\
        F_{Norm}(CD3+, CD4+) \\
        F_{Norm}(CD19+) \\
        F_{Norm}(CD3-, CD16+, CD56+) \\
\end{pmatrix},
\end{equation}
where the normalization function is as follows:

\begin{equation}
    F_{Norm} = \frac{LabVal_{Abs} - ref_{low}}{ref_{high} - ref_{low}},
\end{equation}
where $F_{Norm}$ is applied over an individual absolute lab value, $LabVal_{Abs}$, with reference ranges, $ref_{low}$ and $ref_{high}$, dependent upon the hospital where laboratory testing was performed.

Additionally every vector can be mapped to the disease label and the age of the patient at the time of data collection, although the disease label is blinded when running the clustering algorithms. The vectors are organized into the age subsets, which are then individually analyzed using the unsupervised clustering algorithms.

\subsubsection*{\textbf{Clustering}}

All clinical data is organized in 5-dimensional representation, with the values of each vector component being the normalized values from the lab tests. We implemented five clustering algorithms including: DBSCAN, Hierarchal DBSCAN (HDBSCAN), K-means , agglomerative, and K-modes clustering. Every algorithm is tested for many combinations of hyperparameters which are used to calculate metrics through grid search. In addition to the principal component analysis, the following hyperparameters were optimized: epsilon distance for DBSCAN and HDBSCAN, number of clusters for K-means, K-modes and agglomerative clustering.  

\begin{figure}[htbp]
    \centering
    \begin{subfigure}[]{0.5\textwidth}
        \centering
        \includegraphics[width=\textwidth]{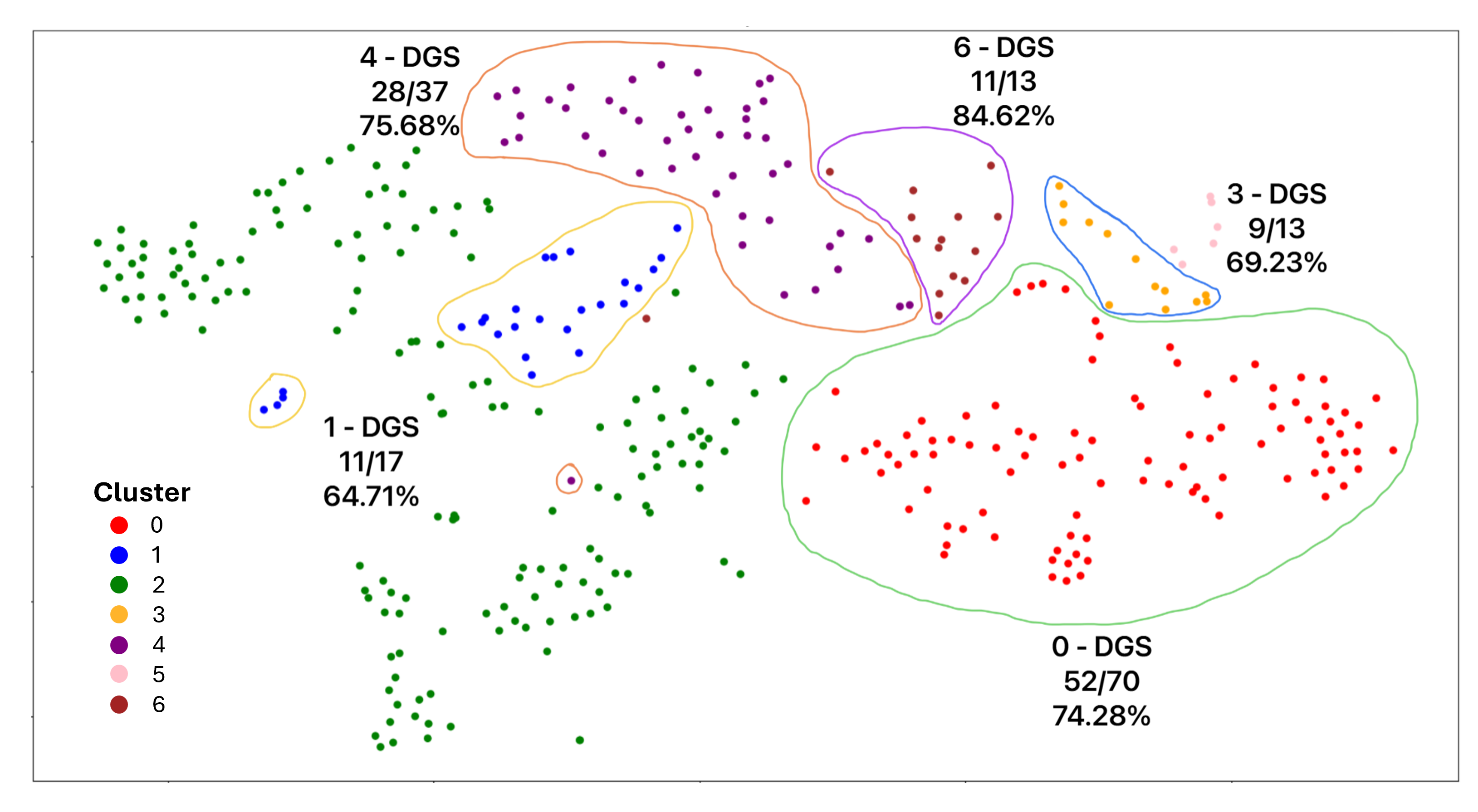}
        \caption{Agglomerative clustering on 0 to 1 year old age group.}
        \label{fig:3a}
    \end{subfigure}
    ~ 
    
    \begin{subfigure}[]{0.5\textwidth}
        \centering
        \includegraphics[width=\textwidth]{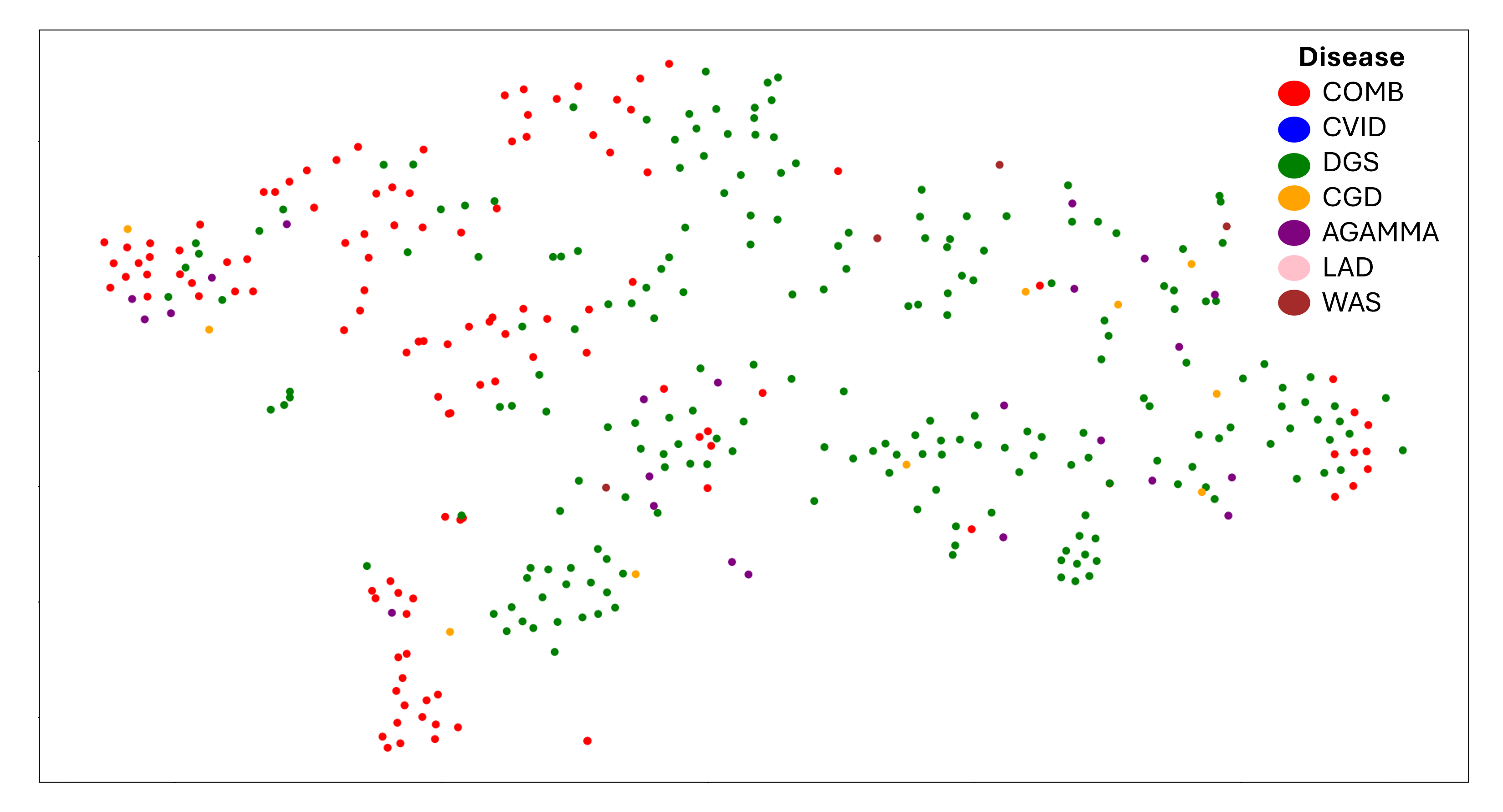}
        \caption{Original patient labels for 0 to 1 year old age-group.}
        \label{fig:3b}
    \end{subfigure}
    \caption{High composite score clusters for 0-1 year old agglomerative clustering experiment. Each patient data point is a 5-dimensional vector reduced through PCA followed by further reduction
into 2 dimensions via t-SNE algorithm to project onto a 2D space. The illustrations show the dominant disease group in the  enclosed cluster and it's fractional proportions with the actual count.}
    \label{fig:combined}
\end{figure}
To evaluate the generated clusters, we calculate the homogeneity score, completeness score, V-Measure, adjusted random index, and adjusted mutual information using known labels on clusters with unknown labels. Additionally, to evaluate the internal information of every cluster, we calculate the Davies Bouldin and the Silhouette scores.

Further, to evaluate the combined performance across metrics, we calculate a composite score that we define as follows:

\begin{equation}
    CS = (\frac{1}{1+DB}) + {\sum_{i=1}^{n} s_i}, 
\end{equation}
where, $CS$ is the Composite score for given combination of hyperparameters.; $s_i\in$ [Homogeneity, Completeness, V-Measure, Adjusted Rand Index, Adjusted Mutual Information, Silhouette Coefficient]; $DB$ - Davies Bouldin (DB) score. The DB score is transformed in this manner since the value is inversely proportional to the performance. Also, the score is offset by 1 because the range of DB score is $[0,\infty)$.

The resultant score table for hyperparameter combinations is sorted by composite score. The top 5 combinations were selected and the median hyper-parameters were calculated from them. For integral values, these calculated hyper-parameters were rounded up to the closest integer. 


\section{Experimental Results}

A total of 5,986 patient-encounter lab data across 2,174 patients with IEIs based on health record diagnostic codes were included in our study. Five clustering algorithms (DBSCAN, HDBSCAN, Hierarchical, K-means, K-modes) were analyzed with a total of 524 hyperparameter combinations across six age groups (DBSCAN 160; HDBSCAN 160; Hierarchical 68; K-means 68; K-modes 68), bringing the total number of experiments to 3,144. Experiments were performed on data sub-grouped by age. All 5 algorithms were tested using principal component analysis (PCA) with components ranging from 2 to 5. For DBSCAN and HDBSCAN algorithms, their epsilon distances were varied from 0.005 to 0.2 with a step size of 0.005. The results are presented in Table 2. 

\begin{table*}[!t]
    \centering
    \caption{A summary of results showing the optimal hyperparameter, Silhouette coefficient, and composite scores for each clustering algorithm using MICE imputation. ED: epsilon distance, SC: Silhouette Coefficient, CS: Composite Score, NC: number of clusters.}
    \scalebox{0.9}{
    \begin{tabular}{|c|cccc|cccc|cccc|cccc|cccc|}
        \hline
        \textbf{Age} 
        & \multicolumn{4}{|c|}{\textbf{DBSCAN}} 
        & \multicolumn{4}{|c|}{\textbf{HDBSCAN}} 
        & \multicolumn{4}{|c|}{\textbf{K-Means}} 
        & \multicolumn{4}{|c|}{\textbf{Agglom}} 
        & \multicolumn{4}{|c|}{\textbf{K-Modes}} \\
        
        \cline{2-21}
        \textbf{Group} 
        & \textit{PCA} & \textit{ED} & \textit{CS} & \textit{SC} 
        & \textit{PCA} & \textit{ED} & \textit{CS} & \textit{SC}
        & \textit{PCA} & \textit{NC} & \textit{CS} & \textit{SC} 
        & \textit{PCA} & \textit{NC} & \textit{CS} & \textit{SC}
        & \textit{PCA} & \textit{NC} & \textit{CS} & \textit{SC} \\ \hline
        
        \hline
        0-1 y 
        & 2 & 0.185 & 2.22 & 0.19 
        & 2 & 0.175 & 2.52 & 0.2
        & 3 & 7 & 2.75 & 0.29 
        & 5 & 7 & 2.37 & 0.26 
        & 4 & 5 & 0.23 & -0.42 \\
        \hline
        1-2 y 
        & 2 & 0.185 & 2.52 & 0.22 
        & 2 & 0.175 & 2.39 & 0.18 
        & 4 & 5 & 2.94 & 0.36 
        & 3 & 7 & 2.83 & 0.31 
        & 3 & 4 & 0.28 & -0.4
        \\
        \hline
        2-5 y 
        & 2 & 0.185 & 2.15 & 0.22 
        & 2 & 0.18 & 2.29 & 0.34 
        & 4 & 8 & 2.36 & 0.35 
        & 2 & 7 & 2.03 & 0.32 
        & 4 & 4 & 0.21 & -0.42 \\
        \hline
        5-10 y 
        & 2 & 0.145 & 1.73 & 0.08 
        & 2 & 0.18 & 1.96 & 0.04 
        & 5 & 7 & 1.94 & 0.27 
        & 5 & 8 & 2.06 & 0.22 
        & 4 & 5 & 0.45 & -0.33 \\
        \hline
        10-16 y 
        & 2 & 0.185 & 2.26 & 0.26 
        & 3 & 0.095 & 2.29 & 0.31 
        & 4 & 5 & 1.84 & 0.29 
        & 4 & 5 & 1.56 & 0.26 
        & 4 & 4 & 0.23 & -0.48 \\
        \hline
        16+ y 
        & 2 & 0.18 & 2.49 & 0.42 
        & 2 & 0.19 & 2.4 & 0.45 
        & 4 & 5 & 2.01 & 0.96 
        & 4 & 5 & 1.79 & 0.78 
        & 3 & 5 & 0.07 & -0.54 \\
        \hline
    \end{tabular}}
\end{table*}

Each patient data sample, a vector of 5 normalized lab values, went through dimensionality reduction through PCA followed by further reduction into 2 dimensions via t-SNE algorithm for visualization. The visualization via t-SNE demonstrates the original patients with their disease labels as well as patient groupings after clustering (Figure 3). Matching of blinded disease labels back to each patient after clustering is demonstrated in the text overlay showing the predominant disease group in each cluster.

An example experiment based on composite score and clinical validation is demonstrated in Figure 3. Figure 3a visualizes agglomerative clustering with PCA set to 5 and cluster number set to 7 for lab data collected when patients were between 0 to 1 years old. Out of the 7 clusters formed, all but one were predominated by data for DiGeorge Syndrome patients. Circled are clusters where 60-90 percent of lab samples in the cluster belong to the DGS patients. 

For clinical validation of example clusters in Figure 3a, we evaluated the most frequent diagnostic codes for the predominant disease patients in each cluster. This allowed us to conduct an initial assessment in the differences in diagnoses/symptoms across patients with same disease, but assigned to different clusters. We acknowledge the ICD codes can be a noisy proxy for clinical significance and requires further validation with lab values, genetics, and holistic chart review. Large ICD code repositories, nonetheless provide an informative lens into recurring patient conditions which we took into account with analysis of all ICD codes that were recorded in patient charts in addition to a review of codes that occurred multiple times in an individual chart with at least 90 days in between recordings. 

In looking at the top 20 ranked diagnostic (ICD) codes across each cluster (excluding IEI defining codes), there was a difference in the most frequent codes between DGS clusters 0, 3, 4, and 6. For DGS patients in cluster 0, the highest frequency ICD codes were related to acute gastrointestinal and feeding problems 
while DGS patients in cluster 3 had more ICD codes related to upper airway pathology and infection
DGS clusters 4 and 6 had a predominance of cardiac related ICD codes although cluster 6 included codes related to aortic anomalies 
which did not appear in the other top 20 code lists. Subsequent clinical validation will focus on using smaller instutition specific IEI datasets including genetyping and additional laboratory value distribution between clusters.



Overall, the best performing clusters by composite score and clinical validation were fixed cluster algorithms, K-means (CS 2.94) and Agglomerative (CS 2.83). 
Despite the best performing clusters by sillhouette and composite score, the density based algorithms produced low number of clusters relative to the diseases examined or clusters of very small sizes based on total number of data points included. The majority of clusters formed across all algorithms were for AGAMMA and DGS disease groups, the two largest population sets included in this pilot study. 


\section{Discussion}
We used laboratory data for immune system function and demonstrated that highly performing clustering experiments are able to group patients together with the same disease and further subdivide those patients into subgroups with differences in diagnostic and symptom code frequencies. 
Overall, we are able to demonstrate that systematic clustering of large scale rare disease data has the potential to identify new clinically relevant patterns. We also identified the necessary statistical and machine learning considerations with data curation and automated analysis of clustering on pediatric rare diseases.

This clustering pipeline has the potential to further subdivide patients with one known disease. Figure 3. demonstrates the example of DGS, a disease with a wide range of presentations including heart defects, facial abnormalities, and hormone imbalances. As the combination of genetic variants and environmental factors lead to new disease presentations, there will be needs for new approaches to treatment. For example, DGS clusters 1, 4, and 6 in Figure 3 all have a cardiac signature but differences in frequencies of specific cardiac abnormalities. Next steps for this validation include evaluating the power and statistical thresholds warranting further investigation of how clusters differ in clinical course or genetic testing that may impact therapy and screening tests needed throughout childhood. Clustering experiments are an opportunity to identify new presentations and patterns of rare diseases. 

In addition to identifying rare disease subgroups and patterns, the second focus of this study is rooted in clustering considerations that need to be taken into account for complex clinical data. This proposed algorithm and data curation highlight that age based data subdivision and age based analysis are critical in pediatric and immunologic disease evaluation. Additionally, this highlights how clustering results can be skewed by unbalanced sample populations, a very common data concern with rare diseases. Data missingness is a very common problem across EHR data warehouses and the approach to using disease and age subgroup specific imputation via an approach like MICE allows for preservation of large datasets and disease differences. 

The systematic evaluation of clustering algorithms in this study demonstrated that automated scoring for further cluster evaluation needs to take into account both well known scoring metrics for clustering (i.e. Silhouette score) but also clinically related metrics based on the hypothesis for clustering; Multiple clusters per disease group with a predominant disease beyond random distribution indicated a subphenotype warranting further evaluation.

Overall, we highlight the benefits of unsupervised ML to identify new patterns within clinical data and the critical evaluation that needs to happen with highly complex data housed in electronic health records. The progression of this work will include clustering models with mixed clinical data types followed by classification models on controls versus test patients to establish a risk and prediction model for patients with IEI.

While our clustering pipeline successfully identified clinically relevant patterns, several methodological and data-related limitations should be acknowledged. First, we performed clustering across multiple IEI disease types despite significant imbalances in population sizes. Consequently, the resulting clusters showed a higher prevalence of a few dominant IEIs with larger cohorts, such as DGS and AGAMMA. This pattern is expected in real-world immunodeficiency populations where certain diagnoses are inherently overrepresented. Importantly, we observed that patients with the same disease type preferentially clustered together, supporting the conclusion that the model captures biologically meaningful structure rather than solely reflecting sample size effects. These findings likely represent a first-order stratification driven by major disease classes, while finer subphenotypic variation may be masked in this global analysis. To address this limitation, subsequent work is applying disease-specific clustering approaches to reduce prevalence-driven effects and enable higher-resolution subphenotyping within individual diagnostic groups. Additional strategies, such as cohort balancing or weighted clustering metrics, will also be explored in full population analyses.

Also, we used five normalized lymphocyte subset counts (relative to institution-specific reference ranges) to construct patient representation vectors as a simple and biologically informed preliminary representation. This approach enabled standardized comparison across heterogeneous data sources while minimizing missingness and inter-laboratory variability. Given the central role of lymphocyte subsets in immunodeficiency phenotyping, this provided a clinically interpretable starting point for clustering. However, we recognize that this representation does not fully capture the phenotypic complexity of IEIs. Ongoing work is incorporating expanded laboratory, ICD-based, and demographic feature sets, alongside genetic data, to evaluate their impact on clustering stability and subphenotype resolution. Together, these limitations highlight the necessity of iterative pipeline refinement as larger, more comprehensive EHR datasets become available for validation.

\section{Conclusions}
This study demonstrates ML clustering on IEI clinical laboratory data and how a data processing, cluster interpretation, and clinical validation pipeline must incorporate computational and disease based knowledge for iterative pattern analysis. 
This approach has the potential to identify new disease features and subgroups, or sub-phenotypes, which may impact earlier diagnosis and targeted treatments in IEI.  

\section*{Acknowledgment}
We would like to acknowledge the Immune Deficiency Foundation for funding this project, Dr. Anita Patel and Dr. Hiroki Morizono for their assistance with the Cerner Real World Data environment. 

\section*{Ethics and Compliance}
All the Data used for this study was accessed through and is hosted on the Cerner Real World Data (CRWD) [2], which has a massive amount of de-identified patient data collected from hospital sites across USA. CRWD contains de-identified data collected under appropriate data use agreements.

\vspace{12pt}
\color{red}

\end{document}